\documentclass[conference]{IEEEtran}
\IEEEoverridecommandlockouts

\usepackage{cite}

\usepackage{microtype}
\usepackage{graphicx}
\usepackage{subcaption}

\usepackage[utf8]{inputenc} 
\usepackage[T1]{fontenc}    
\usepackage{hyperref}       
\usepackage{url}            
\usepackage{booktabs}       
\usepackage{amsfonts}       
\usepackage{nicefrac}       
\usepackage{microtype}      
\usepackage{xcolor}         
\usepackage{algorithm}
\usepackage{algorithmicx}
\usepackage{algpseudocode} 
\usepackage{epstopdf}
\usepackage[english]{babel}
\usepackage[square,numbers]{natbib}
\usepackage{comment}
\usepackage{balance}
\usepackage{enumitem}
\usepackage{xcolor}

\usepackage{grffile}
\usepackage{tikz}
\usepackage{float}

\usepackage{hyperref}

\usepackage{amsmath}
\usepackage{amssymb}
\usepackage{mathtools}
\usepackage{amsthm}
\usepackage{comment}

\usepackage[labelfont=bf]{caption}
\newcommand{\R}      {\mathbb{R}}

\renewcommand*{\top}{{\mkern-1.5mu\mathsf{T}}}

\newcommand{\X} {X} 
\newcommand{\XX}{Y} 

\newcommand{\myBullet}[1]		{\rule{4pt}{4pt}}

\theoremstyle{plain}

\theoremstyle{definition}

\theoremstyle{remark}

\setcitestyle{square,numbers,citesep={,},aysep={,},yysep={;}}
\renewcommand{\citet}[1] {\cite{#1}}

\begin{document}

\title{A framework for paired-sample hypothesis testing for high-dimensional data\\

\thanks{$\square$~The authors acknowledge the support from the Industrial Data Analytics and Machine Learning Chair hosted at ENS Paris-Saclay%
. Correspondence to: \\
\{ioannis.bargiotas;\,argyris.kalogeratos\}@ens-paris-saclay.fr.}}

\author{\IEEEauthorblockN{Ioannis Bargiotas\qquad Argyris Kalogeratos\qquad Nicolas Vayatis}
\IEEEauthorblockA{\textit{Centre Borelli, ENS Paris-Saclay}
\\
Gif-sur-Yvette, France}
}

\maketitle

\begin{abstract}
The standard paired-sample testing approach in the multidimensional setting applies multiple univariate tests on the individual features, followed by $p$-value adjustments. %
Such an approach suffers when the data carry numerous features. A number of studies have shown that classification accuracy can be seen as a proxy for two-sample testing. However, neither theoretical foundations nor practical recipes have been proposed so far on how this strategy could be extended to multidimensional paired-sample testing. %
In this work, we put forward the idea that scoring functions can be produced by the decision rules defined by the perpendicular bisecting hyperplanes of the line segments connecting each pair of instances. Then, the optimal scoring function can be obtained by the pseudomedian of those rules, which we estimate by extending naturally the Hodges-Lehmann estimator. %
We accordingly propose a framework of a two-step testing procedure. First, we estimate the bisecting hyperplanes for each pair of instances and an aggregated rule derived through the Hodges-Lehmann estimator. The paired samples are scored by this aggregated rule to produce a unidimensional representation. %
Second, we perform a Wilcoxon signed-rank test on the obtained representation. Our experiments indicate that our approach has substantial performance gains in testing accuracy compared to the traditional multivariate and multiple testing, while at the same time estimates each feature's contribution to the final result.
\end{abstract}


\section{Introduction}\label{sec:intro}

In many situations, it occurs that the data instances of two samples are \emph{paired}, as they correspond to two measurements of the same population of subjects. %
The problem of testing the homogeneity of \emph{two paired samples} arises in various applications, ranging from medicine to finance. Especially in medicine, comparing a cohort \emph{before} and \emph{after} a specific treatment is very common. Scientists are interested in knowing if there is any statistical difference between the paired samples, and they investigate that either by multivariate hypothesis testing (e.g.~using the multivariate Hotelling $T^2$-test, HT2 for short) or by relying on a \emph{multiple testing} (MT) strategy that performs a series of univariate tests. For instance, one can apply the MT strategy using the paired $t$-test, which comes with a normality assumption about the data. Another possibility is to use the well-known one-sample Wilcoxon signed-rank test (WSR), which is based on linear rank statistics. WSR is one of the most widely-used nonparametric approach in paired-sample hypothesis testing (a resource for several univariate tests can be found in \citet{hollander2013nonparametric}). Rank statistics have been introduced to avoid making assumptions about the distribution of a data sample under the null hypothesis. %
However, the MT strategy has raised a well-known scientific debate concerning mainly the significant increase of Type I error (false positive rate). Therefore, MT is usually followed by $p$-value adjustments that aim to control this increase \citet{feise2002multiple}. Among the various proposed $p$-value adjustments, Bonferroni correction is the one used in the vast majority of studies, %
as it is simple to comprehend and easy to calculate%
, although it is slightly less powerful than others \citet{HochbergHommel2014}.

Besides the natural interest in Statistics, the Machine Learning community %
has shown an increasing interest in hypothesis testing. For instance, in the case of two-sample testing, they have proposed meaningful solutions of how to learn an optimal decision function (i.e. a classifier) with which the dataset can be optimally scored and -in a second step- tested for any significant change between the samples \citet{kim2021classification,rosenblatt2021better,lopez2016revisiting,hediger2022use}. %
Many studies have stressed the importance of choosing a suitable classification criterion for this purpose. For example, the fact that the empirical AUC can be viewed as the extension of Mann-Whitney $U$-statistic in the multivariate setup \citet{cortes2003auc,clemenccon2008empirical}, brought the conclusion about the superiority of maximizing the area under the ROC curve (AUC) criterion in non-parametric two-sample testing. This result led to a number of algorithms for two-sample testing for multidimensional data, based on the AUC maximization \cite{clemenccon2009auc,bargiotas2021revealing}. %

Despite the excessive need for paired-sample analysis, neither theoretical foundations nor practical recipes have been proposed for how such an optimal decision function can be defined for paired data samples. %
The general goal of this work is to %
rely on pairs of instances for defining an optimal scoring function that will allow the scoring of the dataset, the application of paired-sample testing, and the identification of data features that have significantly changed between the two samples. %
To this end, we present a framework for multidimensional paired sample testing, which is inspired by the testing procedures introduced in prior works for the two-sample setting \citet{clemenccon2009auc,bargiotas2021revealing,hediger2022use}. %
At the core of our approach lies the definition we introduce for the \emph{pseudomedian hyperplane}, formulated by the Hodges-Lehmann estimator, which essentially allows us to see the paired-sample testing problem from the classification viewpoint.

\smallskip
\noindent{\textbf{Motivational example}}
\\
When researchers want to compare a cohort before and after a specific medical treatment using multiple features, they can employ already proposed multivariate hypothesis tests. Despite their usefulness, such %
tests often have limitations %
in several dimensions \citet{bai1996effect}, or they might exhibit inadequate power when dealing with slightly difficult cases, such as heavy-tailed multivariate data \citet{purdom2005error}. For this reason, the use of multivariate tests, e.g.~the HT2 test and its variants, is less common. Researchers usually rely on an MT procedure to find those features that change significantly after the treatment. Nevertheless, the %
plethora of acquisition modalities and computed features challenge the hypothesis testing when following this approach. %
Many works either fail to correctly use MT by adjusting the $p$-values \citet{williams1997low}, or they do not proceed to such adjustment because it is criticized for a significant increase in Type II error (false negative rate) \citet{feise2002multiple,Perneger1998}. %
For that reason, several biostatisticians recommend disclosing all the results (significant or non-significant) of the conducted analysis. The violation of this recommendation, the habitual misuse of those tests \cite{Thiese2015}, the little agreement on when or how to adjust the $p$-values, combined with the relatively small available cohorts, may lead to false conclusions, lack of reproducibility, and, as a consequence, a significant delay to reach scientific consensus. Such problems led certain scientific communities to redefine statistical significance and to embrace debatable guidelines \citet{benjamin2018redefine}. 

\smallskip
\noindent\textbf{Contribution}
\\
This work proposes a framework for multivariate paired-sample hypothesis testing by %
aggregating information extracted independently from each pair of instances. %
For each pair, we compute a \emph{perpendicular bisecting hyperplane}, which is perpendicular to the line segment connecting the two instances and passes from the segment's midpoint. %
By definition, any point on such a hyperplane is equidistant to the two associated instances, and also separates the data space in two parts, similar to what the midpoint does in the unidimensional setting. %
We propose an aggregation of those %
hyperplanes in the vain of a \emph{pseudomedian} %
estimated by the Hodges-Lehmann estimator, which is %
directly related to the Wilcoxon signed rank (WSR) test, where it has been proven to be a consistent estimator for the unidimensional case. We call our test \emph{Multivariate WSR} (MWSR), since its principle can be seen as a multivariate extension of the WSR test, where the Hodges-Lehmann pseudomedian hyperplane provides a scoring of the data.
Although the perpendicular bisecting hyperplanes provide `naive' linear decision rules that separate the two samples, yet their pseudomedian can be an adequate decision rule that can be used for scoring and ranking the instances. %
Our approach avoids the $p$-value adjustment, and the contribution of every feature to the final result can be measured using the
linear coefficients of the pseudomedian hyperplane. %

Finally, we provide empirical evidence that our test constitutes an important alternative to multivariate test or MT (especially for marginally separated cases) as it leads to clearly improved performance. %

\section{Background}\label{sec:background}

\subsection{Wilcoxon Signed Rank test (WSR)}\label{sec:Wilcoxon}

Suppose we are in the univariate setting, the two paired samples are denoted by ${\X,\XX}\in\mathbb{R}$, and $(\X_i,\XX_i)$ is the $i$-th out of the $N$ pairs of data instances that corresponds to two measurements of the same subject, the $i$-th of the population. The WSR test has been designed to test whether a population is symmetric around a given median. In the context of the two-sample testing, the WSR test is used to study the population $Z = \{Z_1,..., Z_N\}$ comprising the paired differences $Z_i = \XX_i - \X_i$, for $i = 1, ... ,N$. %

\smallskip
\noindent\textbf{Assumptions}
\begin{enumerate}[topsep=0pt,itemsep=0ex,partopsep=1ex,parsep=0.5ex,leftmargin=3ex]
\item The differences $Z_1, ... , Z_N$ are independent and identically distributed (i.i.d.), i.e.~$Z_i\sim F$, where $F$ represents their continuous probability density function.

\item %
$F$ should be a symmetric distribution around its median $\theta$, which is usually referred to as \emph{effect size} (e.g.~the effect size of a certain medical procedure) \citet{hollander2013nonparametric}. Formally, this suggests that: 
$\theta$: $F(\theta + x) = F(\theta - x)$, $\forall x\in\mathbb{R}$.
\end{enumerate}

\smallskip
\noindent\textbf{Null and alternative hypotheses}
\\
The type of change the WSR detects is a distribution shift from $X$ to $Y$ by a quantity $\theta$, which is defined as the median of the differences contained in $Z$. Thinking in regard to the pairwise differences $Z_i$, essentially converts the paired-sample testing problem to one-sample testing over the population $Z$. Consequently, the null and the alternative hypotheses of the WSR paired-sample test can be derived from those of the one-sample test. %
The two-sided WSR test \cite{wilcoxon1945} considers:
\begin{equation}
\begin{aligned} \label{eq:wsr_hypotheses}
\textup{H}_0 : \theta = 0,\ \ & \text{the } Z_i \text{'s are symmetric around } \theta = 0;\\
\textup{H}_1 : \theta \neq 0, \ \ & \text{the } Z_i \text{'s are symmetric around } \theta \neq 0.
\end{aligned}%
\end{equation}
\noindent \!\! Under the null hypothesis $\textup{H}_0$, there is no difference between the two samples, and therefore the considered treatment or intervention caused a zero shift ($\theta = 0$) in location. This simply asserts that the $Z_i$'s are symmetrically distributed around $0$.

\smallskip
\noindent\textbf{Procedure}
\\
The WSR statistic $T^+_Z$ is calculated from the absolute values $|Z_1|, ..., |Z_N|$ of the differences and their signs. The absolute values are sorted in ascending order, and let $R_i$ denote the rank of $|Z_i|$, $i = 1, ..., N$, in that latter order. Let the positive sign indicator variables $\psi_i$, $i = 1, ... , N,$ where $\psi_i = 1$ if $Z_i > 0$, and $0$ if $Z_i < 0$. The product $\psi_i R_i$ is %
the \emph{positive signed rank} of $Z_i$. %
The WSR statistic $T^+_Z$ is then the sum of the positive signed ranks:
\begin{equation}\label{eq:tag1}
T^+_Z = \sum_{i=1}^{N} \psi_i R_i.
\end{equation}
According to the two-sided version of the test, and at a given significance level $\alpha$,
if the $\X$ and $\XX$ samples differ only by a location shift $\theta$, then the $(Z_i - \theta)$’s have a distribution that is symmetric around $0$. The default procedure is for testing with $\theta = 0$. To test with $\theta = \theta_0 \neq 0$, one needs to subtract a given $\theta_0$ of interest from each $Z_i$ difference to form a modified sample $Z^{(\theta_0)} = \{Z_1 - \theta_0,..., Z_N - \theta_0\}$. %
Then, $T^+_{Z^{(\theta_0)}}$ is computed as the sum of the positive signed ranks for $Z^{(\theta_0)}$.
\subsection{Hodges-Lehmann estimator as effect size}\label{sec:Hodges-Lehmann}

The effect size $\theta$ associated with the WSR statistic $T^+_Z$, is estimated by the \emph{pseudomedian} of the differences, which is in turn estimated by the Hodges-Lehmann estimator $\hat{\theta}$ \citet{HodgesLehmann1963}:
\begin{equation}\label{eq:tag2}
\hat{\theta} = \text{median} \bigg\{ \frac{1}{2}(Z_i + Z_j); \ \forall i \leq j = 1, ... , N \bigg\}.
\end{equation}
In words, this is the median of all the $M = \frac{1}{2}N(N+1)$ pairs of midpoints of the differences. The vector $W_{Z} = [\frac{1}{2}(Z_i + Z_j)]_{ij}$, $\forall i \leq j = 1, ... , N$ is known as the \emph{vector of Walsh averages} \citet{walsh1949some}. Supposing there are no ties and no zeros among the $Z_i$'s, the number of positive Walsh averages ($W^+_Z$) is equal to the WSR statistic $T^+_Z$ \citet{hoyland1965robustness}. %
Without loss of generality, suppose that $F$ satisfies the fact that the median of differences ($\theta$) and the pseudomedian are unique. Then, the Hodges-Lehmann estimator $\hat{\theta}$ is a consistent estimator of the pseudomedian, which can generally be different from the median $\theta$ \citet{HodgesLehmann1963}. However, when $F$ is symmetric, the median and the pseudomedian coincide.

As we mentioned above, %
$\hat{\theta}$ is associated with the WSR test. When $\theta = 0$, then the distribution of the statistic $T^+_Z$ is symmetric around its mean rank $\frac{M}{2}$. An empirical estimator of $\theta$ is the amount $\hat{\theta}$ that should be subtracted from each $Z_i$ so that the value of $T^+_{Z^{(\theta_0)}}$ %
is as close to $\frac{M}{2}$ as possible. Briefly, one needs to calculate the amount that $Z$ should be shifted in order to become a sample with median $0$. Recall that, under Assumption l and 2, each of the ($Z_i - \theta)$'s comes from a population with median that is equal to $0$.

\subsection{Approaches for multivariate paired-sample testing}\label{sec:Hotelling}

Here, we present the basic ideas behind two of the most widely-used approaches for multivariate paired-sample testing, where ${\X,\XX}\in\mathbb{R}^d$, $d>1$. %
One option is to use MT with $p$-value correction. For instance, to test with a target significance level $\alpha$, the Bonferroni correction considers $d$ independent tests, one per dimension, and uses in each of them the `corrected' significance $\frac{\alpha}{d}$ \citet{abdi2007bonferroni}. Another alternative is the multivariate Paired Hotelling $T^2$-test (HT2) \citet{hotelling1992generalization}, which is a generalization of paired $t$-test. Its null hypothesis asserts that the mean vectors $\mu_{X}$, $\mu_{\XX}$, of the population are equal against the alternative hypothesis that these mean vectors are not equal, hence $H_0 : \mu_{X} = \mu_{\XX}$ against $H_1: \mu_{X} \ne \mu_{\XX}$. Same as in the univariate case, the differences between the paired instances are $Z_i = \XX_i - \X_i$, for $i = 1, ..., N$ and the hypothesis testing becomes $H_0 : \mu_{Z} = 0$ against $H_1: \mu_{Z} \ne 0$ where $\mu_{Z} = \mu_{\XX}-\mu_{X}$. %
Under the assumption of normality for $Z$, the HT2 test statistic is given by: %
\begin{equation}\label{eq:T_Hot}
T^2 = N\bar{Z}^{\top}\Sigma_Z^{-1}\bar{Z},
\end{equation}
which is proportional to the distance between $m_Z$ and $0$,  where %
$\bar{Z}$ is the sample mean vector and $\Sigma_Z^{-1}$ is the inverse covariance matrix (or precision matrix). 
\begin{figure}[t]
  \centering
	\begin{subfigure}{0.95\columnwidth}
  \centering
	\includegraphics[scale=0.68,viewport=160 490 450 560,clip]{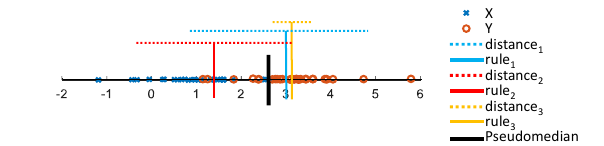}
  \caption{\textbf{Example in $\bf 1$d.}~This is the typical paired-sample setting. We draw as dotted lines the intervals of the differences $Z_1$, $Z_2$, $Z_3$ between the instances of three pairs. The midpoints and the pseudomedian rules are shown as continuous lines vertical to the axis.}
  \label{fig:OneD_to_Two_A}
	\end{subfigure}
	\begin{subfigure}{0.95\columnwidth}
  \centering
	\includegraphics[scale=0.68,viewport=160 315 450 490,clip]{one_to_two_d.pdf}
  \caption{\textbf{Example in $\bf 2$d.}~%
	Here, each distance associated to a pair of instances is shown as a dotted line, while the 1d midpoint rule is now a hyperplane (continuous lines) that is a perpendicular bisector to the line segment connecting the pair, and hence pass from the segment's midpoint.}
  \label{fig:OneD_to_Two_B}
	\end{subfigure}
	\caption{An illustrative example of how we extend the WSR paired-sample hypothesis testing from one dimension (a) to multiple dimensions with the MWSR test (b). We see each pair's midpoint as a decision rule separating the data space, and we generalize this in multiple dimensions by means of perpendicular bisecting hyperplanes. %
	}
	\label{fig:OneD_to_Two}
\end{figure}

\section{Proposed paired-sample testing framework} \label{sec:method}
\noindent\textbf{Testing in multiple dimensions} 
\\
Let us start by establishing that, in the unidimensional case, by adding $\frac{\hat{\theta}}{2}$ to the median of $\X$, we can get what we term as a \emph{pseudomedian decision rule} separating the space in two parts. This is a notion that we generalize to propose our paired-sample framework for multiple dimensions, and the Multivariate WSR test (MWSR). %
 For $d>1$, we deal with the point-cloud of %
${\X,\XX}\in\mathbb{R}^d$, thus %
the Euclidean distance $Z_i = ||Y_i-X_i||_2$ between pairs of instances $(\X_i, \XX_i)$ can be seen as the analogue to the difference $Z_i = \XX_i -  \X_i$ we saw earlier in $1$d. Further, each separating rule associated with a midpoint of the $1$d case, now becomes a ($d\!-\!1$)-dimensional perpendicular bisecting hyperplane: it is perpendicular to the line segment connecting the pair and also passes from its midpoint. %
Each such hyperplane is computed by taking into account only one specific pair of instances, yet it splits the space in two parts, and therefore it can be seen as a \emph{decision rule} that could hopefully classify the data in two parts, the $X$ and the $Y$ part. Fig.\,\ref{fig:OneD_to_Two} shows an insightful visual example: Fig.\,\ref{fig:OneD_to_Two_A} presents the 1d setting, as discussed in the earlier sections: the paired samples lie on a line, each difference $Z_i$ is a distance on that line, and midpoints can be computed for each $(X_i, Y_i)$ pair. %
Fig.\,\ref{fig:OneD_to_Two_B} shows how our approach operates in multiple dimensions. %
The decision rule for a pair $(X_i, Y_i)$ is its perpendicular bisecting hyperplane $C_i : \,w_i^\top x + b_i = 0$, $x\in\mathbb{R}^d$, where $w_i$ are the coefficients of the hyperplane, and $b_i$ is its intercept. With the known elements, it is easy to compute the exact $w_i$ and $b_i$%
; then, $\operatorname{sign}(C_i(x))$ classifies an input datapoint $x$, while %
the signed distance from $x$ to $C_i$ can be seen as a score for $x$ based on this hyperplane.%

In simple terms, we propose to see the scoring function derived from the hyperplane of the $i$-th pair as an analogue to the half of the empirical effect size calculated by this specific pair of instances. Furthermore, the aggregation of all the $N$ rules can provide a more robust scoring function. This aggregation can be computed by directly defining the \emph{pseudomedian decision rule in multiple dimensions}: The linear coefficients $w_i$ and the intercept $b_i$ of each $C_i$ rule are used for estimating the Hodges-Lehmann pseudomedian hyperplane through the $\hat{w}$, using the same procedure as $\hat{\theta}$ (see Sec.\,\ref{sec:Hodges-Lehmann}). %

\smallskip
\noindent\textbf{The two-step MWSR testing procedure}
\\
We compute a linear decision rule for each pair $(X_i, Y_i)$, hence we gather $N$ such models. %
In order to extend the procedure presented in Sec.\,\ref{sec:Hodges-Lehmann} to the general $d$-dimensional setting, we need to define first the \emph{Walsh model-wise average} $W_C = \left[\frac{1}{2}(C_i + C_j)\right]_{ij}$, $\forall i \leq j = 1, ..., N$, as the average of the coefficients and intercepts for pairs of linear models. For the linear decision rules that we employ, this is: 
Having specified how we can pass from averaging instances to averaging model parameters, we can now express the Hodges-Lehmann estimator in the same form as before:
\begin{equation}\label{eq:tag3}
\hat{C^*} = \text{median} \bigg\{ \frac{1}{2} (C_i + C_j); \ \forall i \leq j = 1, ... , N \bigg\},
\end{equation}
which computes the median coefficient-wise. The estimated $\hat{C^*}$  is the empirical optimal scoring function where the final classification scores ($\hat{S_1}$, $\hat{S_2}$) are calculated for $\X$ and $\XX$, respectively. %
Note that the operation of finding $\hat{C^*}$ from a set of $N$ decision rules can be seen as \emph{model aggregation}, which in this case is not a generic one (e.g.~like the \emph{voting} used in Ensemble Learning), but rather a special aggregation, directly linked to the WSR test that we employ next.

The last step amounts to performing a WSR test on $\hat{S_1}$ and $\hat{S_2}$. In addition to the given $p^*$-value and the effect size $\theta^*$, the user also gets informed about the contribution of each dimension to the final result (through the coefficients of the linear model $\hat{C^*}$).

The feature importance can be easily obtained considering that the more orthogonal is a decision rule (hyperplane) to a certain feature axis, the higher the associated coefficient will be. Furthermore, the sign indicates the positive or negative relationship with the response. The procedure of the proposed framework is detailed in Alg.\,\ref{alg:algo}.

\begin{algorithm}[t]\small
\caption{The MWSR paired-sample testing framework %
}\label{alg:cap}
\begin{algorithmic}
\State \textbf{Input: }$\X, \XX \in \R^{N \times d}$ are the $2\cdot N$ paired samples; 
\State \textbf{Output: } $\hat{C}^*$, ($\hat{S}_1^*$, $\hat{S}_2^*$), $p^*\!$-\emph{value}, $\theta^*$, $I^*$
\vspace{1mm}
\hrule
\vspace{2mm}

\State \myBullet~~\textbf{\underline{\emph{First step: Compute a scoring}}}
\For {$i={1,..,N}$}
	\State $C_i \gets $ perpendicular\_bisector$(\X_i,\XX_i)$ %
\EndFor
\State $k \gets 1$; $M \gets {\bf 0}_{N\times N} $
\For {$i={1,..,N}$} \hfill 
\For {$j={i,..,N}$}
		\State $W_{C,k} \gets \frac{1}{2}(C_i + C_j)$ \Comment{the Walsh average of hyperplanes}
		\State $k \gets k+1$
\EndFor
\EndFor
    \State $\hat{C}^* \gets \text{median}(W_C)$ \Comment{the pseudomedian aggregate, see Eq.\,\ref{eq:tag3}}
    \State $\hat{S}_1^*, \hat{S}_2^* \gets \text{get\_scores}(\hat{C}^*(\X,\XX))$ \Comment{classification-based scoring}
		\vspace{2mm}
    \State \myBullet~~\textbf{\underline{%
		\emph{Second step: Paired-sample test over the computed scores}}}
		\State $p^*$\!-\emph{value}, $\theta^*$ $\gets$ WSR$(\hat{S}_1^*,\hat{S}_2^*)$ \Comment{$p$-value and effect size}
    \State $I^* \gets w(\hat{C}^*)$  \Comment{feature importance index}
		\State \Return $\hat{C}^*$, ($\hat{S}_1^*$, $\hat{S}_2^*$), $p^*$\!-\emph{value}, $\theta^*$, $I^*$
\end{algorithmic}
\label{alg:algo}
\end{algorithm}

\begin{figure*}[t]
  \centering
	\begin{subfigure}{0.25\textwidth}
  \centering
	\includegraphics[scale=0.30, viewport=0 2 400 255,clip]{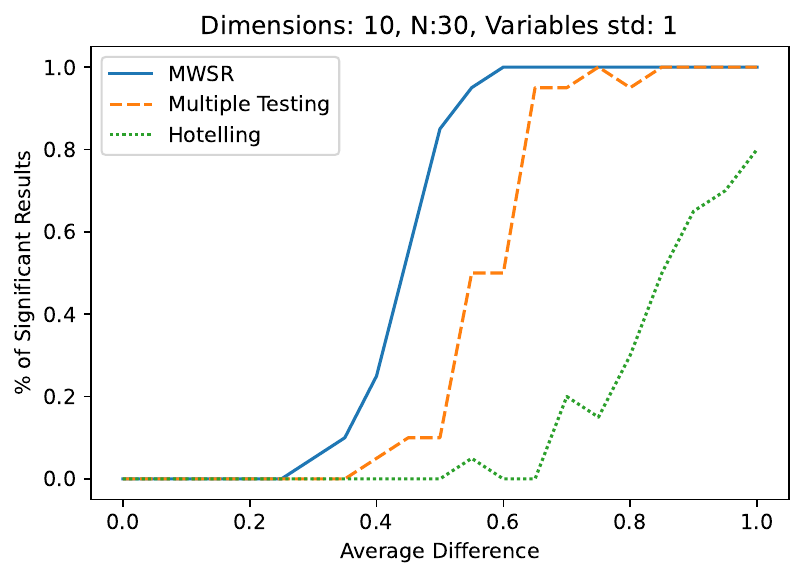}%
  \caption{$d=10$, $N=30$, $std = 1$}
  \label{fig:Sim1030S1}
	\end{subfigure}%
	\begin{subfigure}{0.25\textwidth}
  \centering
	\includegraphics[scale=0.30, viewport=0 2 400 255,clip]{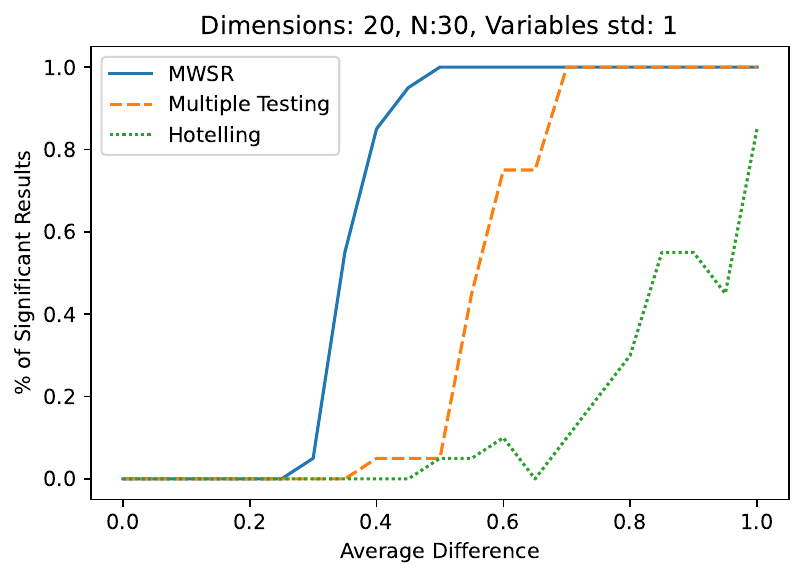}%
  \caption{$d=20$, $N=30$, $std = 1$}
  \label{fig:Sim2030S1}
	\end{subfigure}%
	\begin{subfigure}{0.25\textwidth}
  \centering
	\includegraphics[scale=0.30, viewport=0 2 400 255,clip]{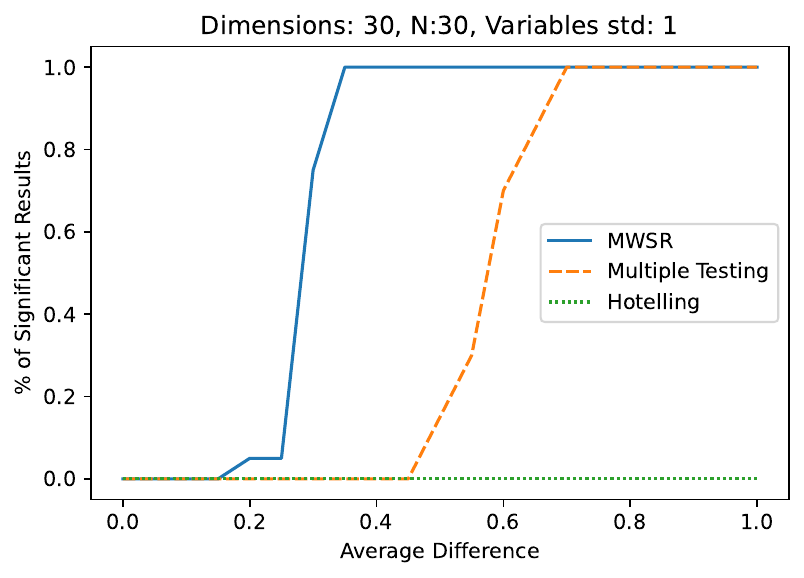}%
  \caption{$d=30$, $N=30$, $std = 1$}
  \label{fig:Sim3030S1}
	\end{subfigure}%
 \begin{subfigure}{0.25\textwidth}
  \centering
	\includegraphics[scale=0.30, viewport=0 2 400 255,clip]{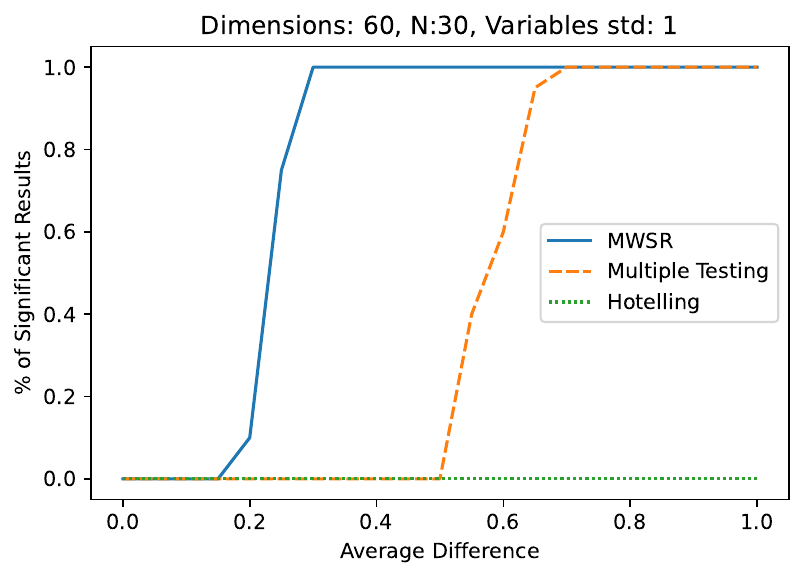}%
  \caption{$d=60$, $N=30$, $std = 1$}
  \label{fig:Sim6030S1}
	\end{subfigure}%
	\\
	\vspace{0.5em}
	\begin{subfigure}{0.25\textwidth}
  \centering
	\includegraphics[scale=0.30, viewport=0 2 400 255,clip]{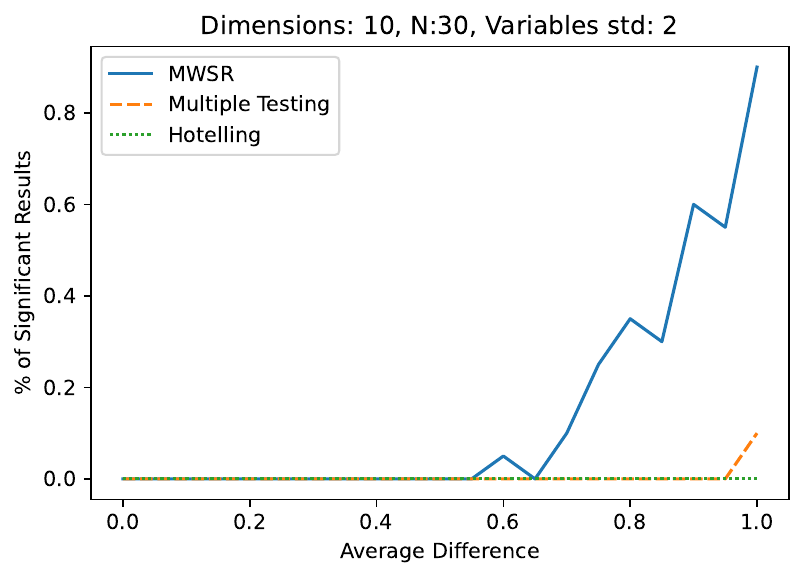}%
  \caption{$d=10$, $N=30$, $std = 2$}
  \label{fig:Sim1030S2}
	\end{subfigure}%
	\begin{subfigure}{0.25\textwidth}
  \centering
	\includegraphics[scale=0.30, viewport=0 2 400 255,clip]{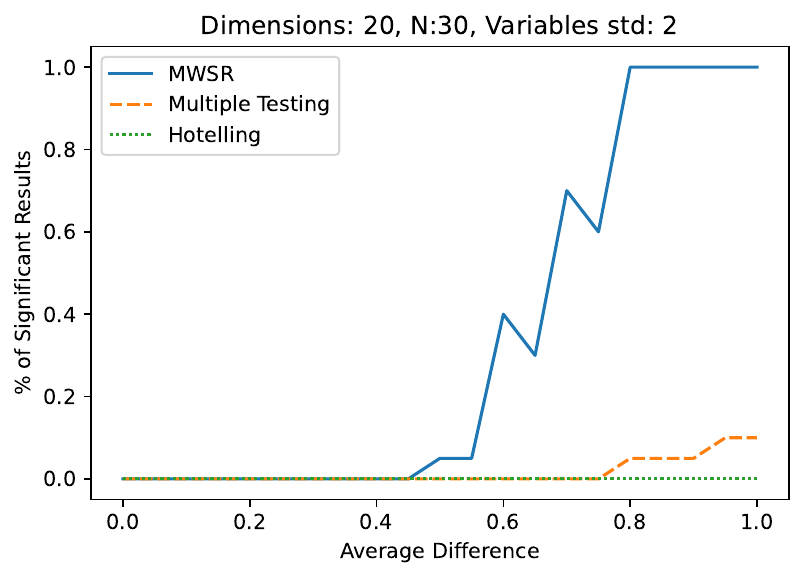}%
  \caption{$d=20$, $N=30$, $std = 2$}
  \label{fig:Sim2030S2}
	\end{subfigure}%
	\begin{subfigure}{0.25\textwidth}
  \centering
	\includegraphics[scale=0.30, viewport=0 2 400 255,clip]{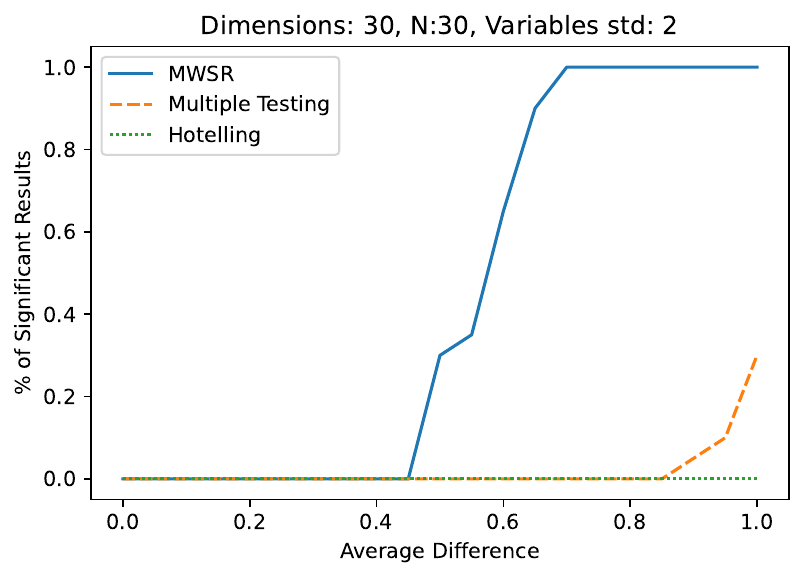}%
  \caption{$d=30$, $N=30$, $std = 2$}
  \label{fig:Sim3030S2}
	\end{subfigure}%
 \begin{subfigure}{0.25\textwidth}
  \centering
	\includegraphics[scale=0.30, viewport=0 2 400 255,clip]{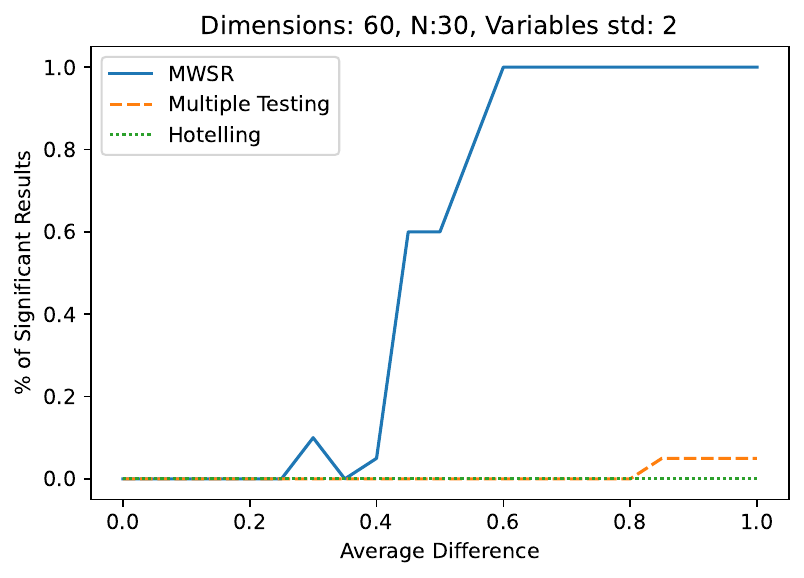}%
  \caption{$d=60$, $N=30$, $std = 2$}
  \label{fig:Sim6030S2}
	\end{subfigure}%
	\vspace{-1mm}
	\caption{The average performance of paired-sample testing approaches in synthetic datasets with $N=30$ pairs of instances coming from two Gaussian distributions, both with either $std = 1$ or $2$ (first and second row, respectively). The performance is presented as a function of the separation distance between the two distributions (average difference on $x$-axis). On the $y$-axis it appears the \% of significant results detected by each method, and on the $x$-axis the progressive difference in the mean value for the $10\%$ of the dimensions ($d$).}
	\label{fig:S1}
\end{figure*}

\begin{figure*}[t]
  \centering
	\begin{subfigure}{0.45\textwidth}
  \centering
	\includegraphics[scale=0.40,viewport=0 188 390 456.8, clip]{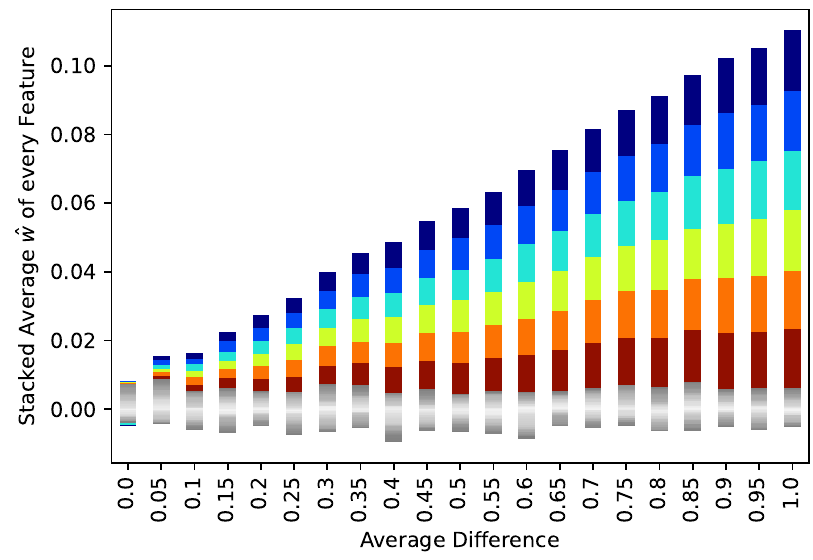}%
  \caption{Average hyperplane coefficients ($\hat{w}$) for each feature ($d=60$) \phantom{......................}}
  \label{fig:Sim6030S1_impSVM}
	\end{subfigure}%
	\ \ \ \ \ 
	\begin{subfigure}{0.45\textwidth}
  \centering
	\ \ \ \ \ \ \ \ \includegraphics[scale=0.40,viewport=0 188 390 456.8,clip]{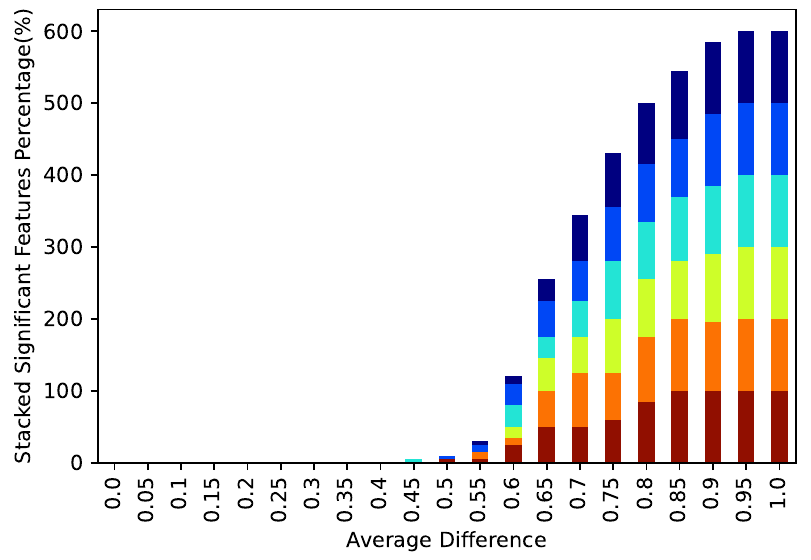}%
	\includegraphics[scale=0.38, viewport=-70 310 60 540,clip]{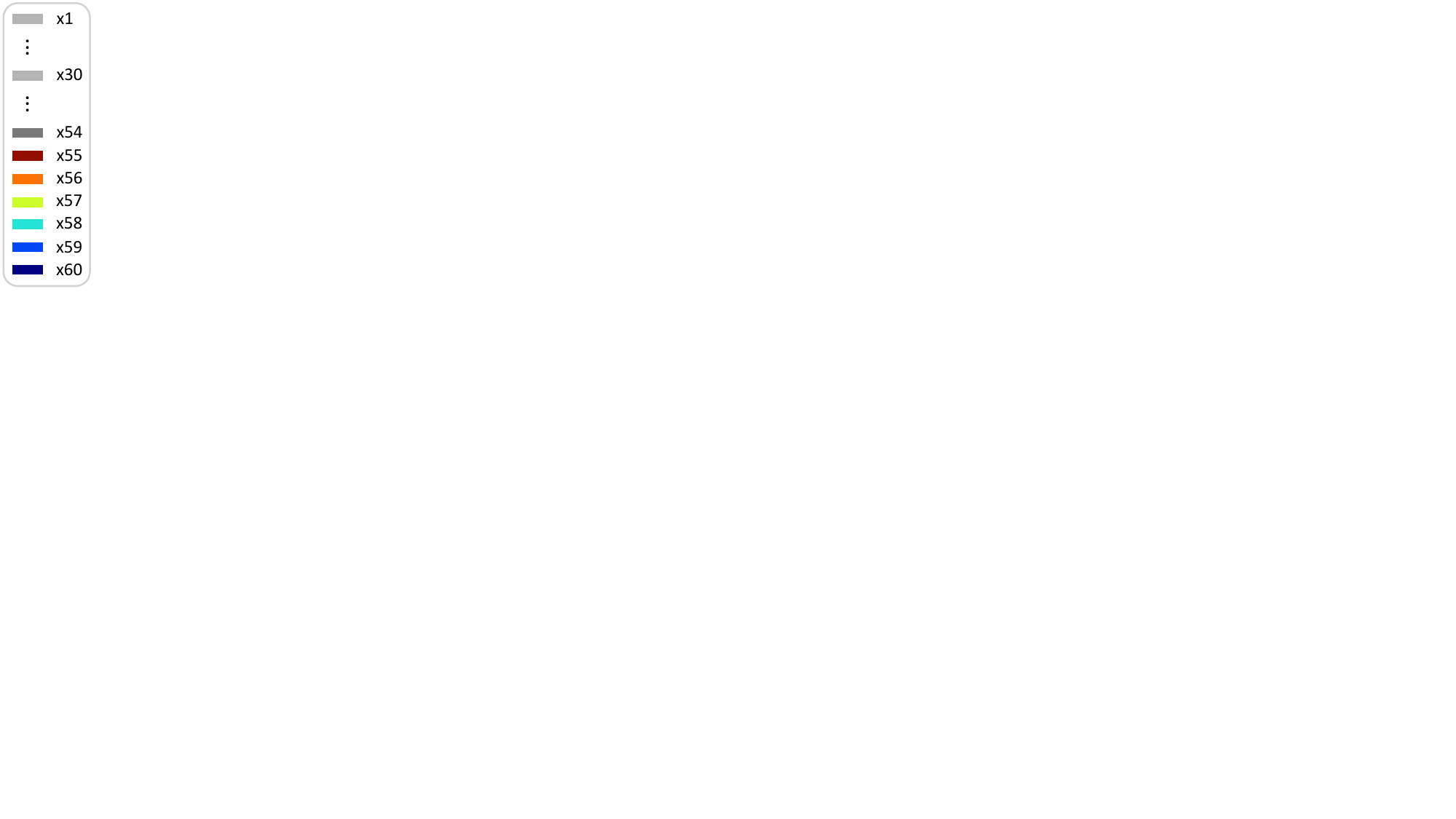}%
  \caption{Percentage of times ($\%$) that features are found significant after MT with Bonferroni correction ($d=60$)}%
  \label{fig:Sim6030S1_impMT}%
	\end{subfigure}%
	\vspace{-1mm}
	\caption{The relative feature importance in the paired-sample testing result of different approaches, %
	and their corresponding average performance for $N=30$ pairs coming from two Gaussian distributions with $std = 1$, in $d=60$ dimensions (corresponds to the case of Fig.\,%
\ref{fig:Sim6030S1}). The performance is presented as a function of the separation distance between the two distributions (average difference on $x$-axis).%
}
	\label{fig:S1_imp}
\end{figure*}
\begin{figure*}[h]
  \centering
	\begin{subfigure}{0.45\textwidth}
  \centering
	\vspace{1mm}
	\includegraphics[scale=0.45, viewport=0 5 453 261,clip]{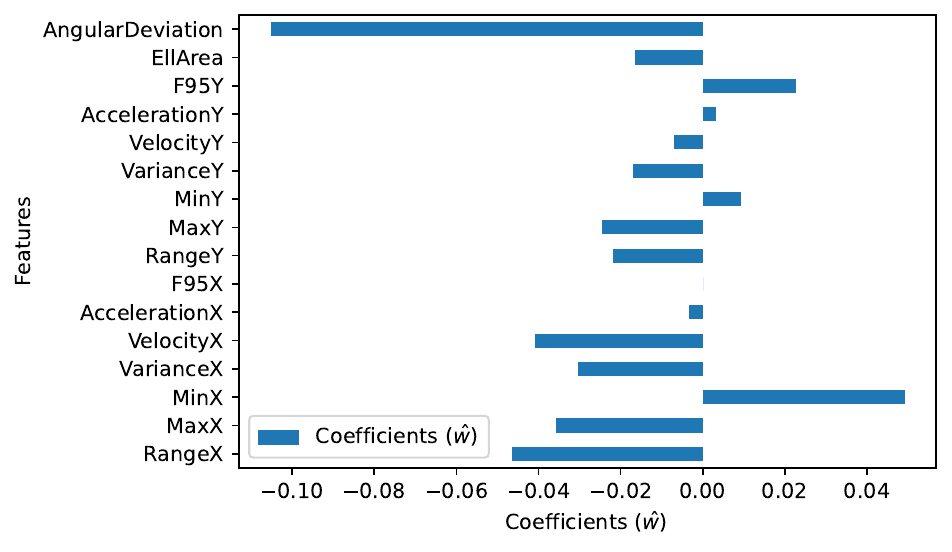}
  \caption{Feature importance ($\hat{w}$) obtained by the MWSR test}
  \label{fig:Real_impSVM}
	\end{subfigure}%
	\begin{subfigure}{0.45\textwidth}
  \centering
	\vspace{-1.5mm}
	\includegraphics[scale=0.452,viewport=0 3 453 261,clip]{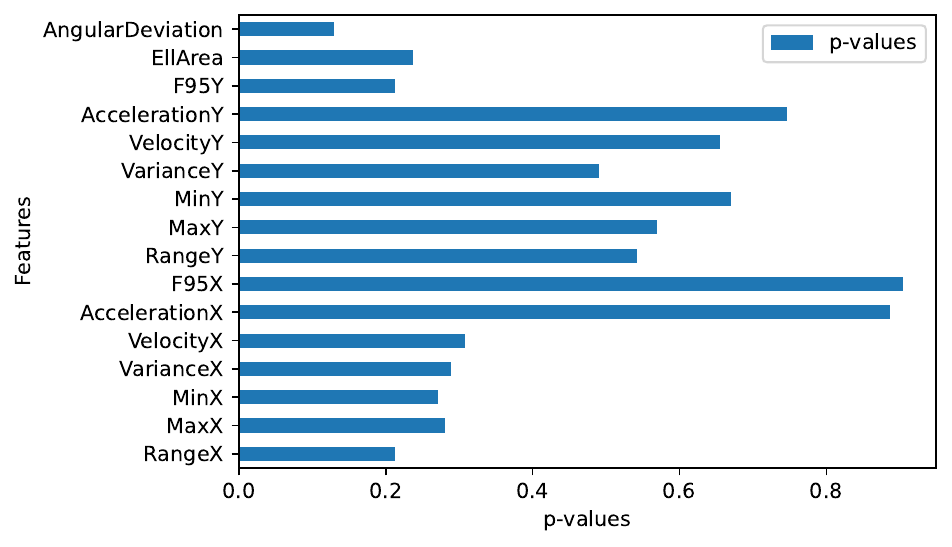}
  \caption{A $p$-value per feature obtained by MT}
  \label{fig:Real_impMT}
	\end{subfigure}
	\caption{The relative importance of each feature %
	as indicated by our approach and the MT on the posturographic dataset ($d=16$ features).}
	\label{fig:Real_imp}
\end{figure*}

\section{Experiments}\label{sec:experiments}

The experiments in this section compare the performance of our testing method, the MWSR test, to the classical multiple testing (MT) procedure with Bonferroni $p$-value adjustment, and the HT2 test. We use several synthetic scenarios and one real clinical dataset. The synthetic scenarios have increasing difficulty and allow us to check the limits of the proposed test. The comparison is two-fold: i) in terms of sensitivity in detecting a difference between the paired samples ($p$-value, effect size), and ii) in terms of estimating accurately the contribution of each feature to the test's outcome, which is helpful for the user to interpret the results.

\smallskip
\noindent\textbf{Synthetic datasets}
\\
Synthetic data are simulated by pairing two samples coming from two Gaussian distributions, with feature-wise correlation:
\begin{equation}\label{eq:corr}
\mathcal{R}_{X^{(k)},Y^{(k)}}={\frac {\operatorname{cov}(X^{(k)},Y^{(k)})}{\sigma _{X^{(k)}}\sigma _{Y^{(k)}}}} = 0.5,
\end{equation}
where $X^{(k)}$, $Y^{(k)}$ are vectors representing the paired samples with only the $k$-th feature, $\operatorname{cov}(\cdot,\cdot)$ is their covariance, and $\sigma_{X^{(k)}}$, $\sigma_{Y^{(k)}}$ are the respective standard deviations. 
To produce the dataset for each scenario, we set:
\begin{itemize}[topsep=0pt,itemsep=0ex,partopsep=1ex,parsep=0.5ex, leftmargin=3ex]
\item a fixed population size, $N=30$ pairs of data instances;
\item the dimensions $d = \{10, 20, 30, 60\}$, mimicking the number of features that a usual study may have;
\item a standard deviation $std = \{1, 2\}$;
\item the first $90\%$ of the dimensions to have no statistical difference between the two samples, and hence to be randomly drawn (separately) from $\mathcal{N}(0, std)$;
\item the last $10\%$ of the dimensions to present the same difference in mean value, hence producing a linear shift. We allow this shift to also increase progressively from $0$ to $1$ to investigate the detection sensitivity of the methods.
\end{itemize}
Given a scenario, we generate $20$ cases and apply all statistical tests. %
We report the average performance, namely the percentage of cases with a significant shift obtained by each statistical test, as a function of the size of the shift (referred to as average difference in the figures). Moreover, we provide results regarding the inferred feature importance. Both elements should be examined jointly to validate the acquired results further.

\smallskip
\noindent\textbf{Real dataset}%
\\
We extend our empirical validation by employing a typical real clinical dataset, with a relatively low population and multiple features. It concerns posturographic assessment for subjects with Parkinson's syndrome (PS). %
This %
dataset, initially used in \citet{bargiotas2019predicting}, includes $30$ subjects (mean age: $79.6 \pm 4.4$ years) from the Neurology department of the HIA, Percy hospital in Clamart, France, who were diagnosed with PS. The subjects underwent a posturography assessment using a force platform (here a Wii Balance Board (Nintendo, Kyoto, Japan)) that captures the trajectory of the center of pressure (CoP) exerted by the entire body over time when an individual stands on them. The assessment comprises two examinations with a $6$-month difference in time, which are the paired samples we use in our experiment. Each time their postural stability was recorded for $25$ seconds while maintaining an upright position on a force platform with eyes open. 

To characterize subjects' postural control, the dataset provides $16$ features that had been previously proposed as indicators of postural stability \citet{quijoux2021review}. In detail: Percentiles ($95\%$ and $5\%$) (cm), Range (cm), Variance (cm$^2$), Mean Instant Velocity (cm/s), Acceleration (cm/s$^2$) and Frequency (Hz) below which $95\%$ of the signal energy is found, for both X-medio-lateral (ML) and Y-antero-posterior (AP) axes, confidence ellipse area (cm$^2$) that covers the $95\%$ of the points of the trajectory and the angular deviation (in degrees$^{\circ}$).

\smallskip
\noindent\textbf{Results and discussion}
\\
In all the synthetic cases in Fig.\,\ref{fig:S1}, the proposed MWSR test is always by far superior to the MT and the multivariate HT2 test in detecting the difference between the paired samples. The MWSR is more sensitive than the compared tests, as it manages to detect much smaller shifts. Moreover, it reports no false positive results in the case of no shift in location. %
Worth noting that when %
$d>N$, the classical HT2 test fails completely %
due to the %
non-existence of the inverse of the sample covariance matrix (see in Fig.\,\ref{fig:S1} the four cases with $d\geq30$). The difference in performance between the MWSR and the MT increases in higher dimensions as well as when the variance is higher (see Fig.~\ref{fig:S1}, e-h), which indicates a robust behavior to that kind %
noise. Note that the MWSR has no particular computational overhead; it runs in less than $0.05$s for the small real dataset, %
and this increases slightly with $N$.

Concerning the feature importance indexes 
obtained by the MWSR, they are effectively recognizing the last $10\%$ of the features that contribute to the distribution shift (see in Fig.~\ref{fig:S1_imp} the features x$55$-x$60$). This is made clear by %
the higher weights (i.e.~larger stacks) assigned to those features. %

In the real dataset, the MWSR test found that the Parkinsonians changed significantly after $6$ months, while the MT did not detect any significant change. 

Despite the above notable advantages, the proposed methodology is designed and limited for linear shifts in location between the paired samples, same as the univariate WSR test. %
Moreover, a point that needs attention is to normalize the data so that the feature importance indexes are better estimated and are easier to interpret. %
The weight assigned to a particular feature is influenced by the other features, particularly when there are correlations among them. %

\section{Conclusions}\label{sec:conclusions}

We presented a sound strategy for paired-sample testing for multidimensional data, which %
outperforms the classical multivariate approaches (Hotelling $T^2$ test, the multiple testing with $p$-value adjustment) in our experiments. %
Our method has two additional advantages: i)~it is generally simple and understandable, %
and ii)~it offers a tool that allows the user to interpret the actual contribution of every feature to the final result. Beyond the illustrative preliminary experiments in real and synthetic datasets that we presented in this paper, we intend to investigate the relative efficiency of this test to other setups. We also plan to investigate theoretically the regime in which this method is qualitatively superior.

\medskip

\balance
{\footnotesize%
\bibliographystyle{icml2022}

}

\end{document}